\documentclass[12pt]{article}

\usepackage{tabularx}
\usepackage{wrapfig}
\usepackage{titlesec}
\usepackage{mathpazo}
\usepackage{ebgaramond}
\usepackage{charter}
\usepackage{bbold}
\usepackage{booktabs}
\usepackage{array}
\usepackage{multirow}
\usepackage{pifont}
\usepackage{xcolor}
\definecolor{myblue}{RGB}{13, 71, 161}   
\definecolor{myred}{RGB}{232, 93, 117}   
\usepackage{tikz}

\usetikzlibrary{positioning, arrows} 
\usetikzlibrary{calc} 
\usepackage{pdflscape}
\usepackage{adjustbox}
\usepackage{subcaption}
\usepackage{colortbl} 
\usepackage{longtable} 

\definecolor{customgray}{gray}{10.5}  
\newcommand{\mycomment}[1]{}
\titleformat{\section}
{\normalfont\Large\bfseries\color{customgray}} 
{\thesection}{1em}{}

\titleformat{\subsection}
{\normalfont\large\bfseries\color{gray}} 
{\thesubsection}{1em}{}
\titleformat{\subsubsection}
{\color{gray}\normalsize\bfseries} 
{\thesubsubsection}{1em}{}
\titlespacing*{\section}{0pt}{1.5ex plus 1ex minus .2ex}{1.5ex plus .2ex}
\titlespacing*{\subsection}{0pt}{1.25ex plus 1ex minus .2ex}{1.25ex plus .2ex}

\usepackage[numbers,sort&compress]{natbib}
\usepackage{times}
\usepackage{tabularray}
\usepackage{heuristica}
\usepackage{wrapfig}
\usepackage{multirow}
\usepackage{booktabs}
\usepackage{algorithm}
\usepackage{algpseudocode}
\usepackage{makecell}

\usepackage[hmargin=1cm]{geometry}

\usepackage{tikz}

\usetikzlibrary{backgrounds}
\usetikzlibrary{patterns}
\usetikzlibrary{bayesnet}
\usetikzlibrary{arrows}
\usepackage{color}

\usetikzlibrary{backgrounds}

\usepackage{array}

\usepackage{amsmath,amsfonts}
\usepackage{amssymb} 
\usepackage{mathtools}

\usepackage{listings}
\usepackage[most]{tcolorbox}

\definecolor{usercolor}{RGB}{255, 108, 108} 
\definecolor{assistantcolor}{RGB}{255, 189, 69} 

\newtcolorbox{userbox}{
    colback=usercolor!5!white,
    colframe=usercolor!75!black,
    width=\linewidth-1cm,
    left=0.5cm,
    right=0.5cm,
    boxrule=0.4mm,
    arc=2mm,
    fonttitle=\bfseries,
    title=Prompt,
    breakable
}

\newtcolorbox{assistantbox}{
    colback=assistantcolor!5!white,
    colframe=assistantcolor!75!black,
    width=\linewidth-1cm,
    right=0.5cm,
    left=0.5cm,
    boxrule=0.4mm,
    arc=2mm,
    fonttitle=\bfseries,
    title=Assistant,
    breakable
}

\lstdefinestyle{pythonstyle}{
    language=Python,
    basicstyle=\ttfamily\small\color{black},           
    keywordstyle=\color{teal}\bfseries,                 
    stringstyle=\color{brown},                          
    commentstyle=\color{gray},                          
    numberstyle=\tiny\color{gray},                      
    frame=single,                                       
    backgroundcolor=\color{gray!5},                     
    breaklines=true,                                    
    showstringspaces=false,                             
    numbers=left,                                       
    xleftmargin=1em,                                    
    rulecolor=\color{black},  
    morecomment=[s][\color{olive!80!black!80}\slshape]{"""}{"""},  
}

\lstdefinestyle{bashstyle}{
    language=bash,
    basicstyle=\ttfamily\small,
    frame=single,
    backgroundcolor=\color{gray!10},
    breaklines=true,
    showstringspaces=false
}

\usepackage{amsmath}
\usepackage{amssymb}
\usepackage{mathtools}
\usepackage{amsthm}
\theoremstyle{plain}

\theoremstyle{definition}

\theoremstyle{remark}

\definecolor{bronze}{RGB}{205, 127, 50}
\definecolor{silver}{RGB}{192, 192, 192}
\definecolor{gold}{RGB}{255, 215, 0}
\usepackage{graphicx}   
\usepackage{amsmath}    
\usepackage{amssymb}    
\usepackage{fancyhdr}   
\usepackage{geometry}   
\usepackage{titlesec}   
\usepackage{xcolor}     
\usepackage{caption}  
\titleformat{\section}{\normalfont\bfseries\Large}{\thesection}{1em}{}
\titlespacing*{\section}{0pt}{*3}{*2}

\usepackage{tcolorbox}

\usepackage{url}      
\usepackage[hidelinks]{hyperref}   

\definecolor{headerblue}{RGB}{250, 245, 245}
\definecolor{titleblack}{RGB}{80, 0, 0} 
\definecolor{abstractblack}{RGB}{80, 0, 0} 

\newtcolorbox{fullbox}{
colback=headerblue,
colframe=white,
width=\textwidth,
boxrule=0pt,
arc=10pt,
outer arc=10pt,
boxsep=10pt,
left=10pt,
right=10pt,
top=10pt,
bottom=10pt
}

\usepackage{todonotes}

\usepackage{pifont}
\def\mark{\ding{108}} 
\begin{document}


\begin{fullbox}
\vspace{0.1em}
\begin{center}
{\Large \textbf{Ark: An Open-source Python-based Framework for Robot Learning}}






\vspace{1em}
\textbf{
Magnus Dierking$^{1}$, 
Christopher E. Mower$^{2,\star}$, 
Sarthak Das$^{2}$, 
Huang Helong$^{2}$,
Jiacheng Qiu$^{1, 2}$,
Cody Reading$^{2}$, 
Wei Chen$^{2, 3}$, 
Huidong Liang$^{2, 4}$, 
Huang Guowei$^{2}$, 
Jan Peters$^{1}$, 
Quan Xingyue$^{2}$,
Jun Wang$^{5,\star}$,
Haitham Bou-Ammar$^{2,5,\star}$}\\
{\small $\ ^1$~Technical University of Darmstadt} 
{\small $^2$~Huawei Noah's Ark} 
{\small $\ ^3$~Imperial College London}\\
{\small $\ ^4$~University of Oxford} 
{\small $\ ^5$~University College London}\\
{\small $^{\star}$~Corresponding authors:}
\texttt{\{christopher.mower, haitham.ammar\}@huawei.com, jun.wang@cs.ucl.ac.uk}
\vspace{0.1em}

\end{center}

\noindent
\textbf{Abstract:} 
Robotics has made remarkable hardware strides-from DARPA's Urban and Robotics Challenges to the first humanoid-robot kickboxing tournament-yet commercial autonomy still lags behind progress in machine learning. A major bottleneck is software: current robot stacks demand steep learning curves, low-level C/C++ expertise, fragmented tooling, and intricate hardware integration, in stark contrast to the Python-centric, well-documented ecosystems that propelled modern AI. We introduce Ark, an open-source, Python-first robotics framework designed to close that gap. Ark presents a Gym-style environment interface that allows users to collect data, preprocess it, and train policies using state-of-the-art imitation-learning algorithms (e.g., ACT, Diffusion Policy) while seamlessly toggling between high-fidelity simulation and physical robots. A lightweight client–server architecture provides networked publisher-subscriber communication, and optional C/C++ bindings ensure real-time performance when needed. Ark ships with reusable modules for control, SLAM, motion planning, system identification, and visualization, along with native ROS interoperability. Comprehensive documentation and case studies-from manipulation to mobile navigation-demonstrate rapid prototyping, effortless hardware swapping, and end-to-end pipelines that rival the convenience of mainstream machine-learning workflows. By unifying robotics and AI practices under a common Python umbrella, Ark lowers entry barriers and accelerates research and commercial deployment of autonomous robots.

\end{fullbox}

\thispagestyle{fancy}






A long-standing goal for the robotics community has been to move robots out of controlled laboratory environments and deploy them commercially in the real world.
There have been several impressive demonstrations over the years of robots operating in challenging environments, such as: 
the 2007 DARPA Urban Challenge, devised to improve self-driving vehicles that can navigate busy city streets and interact safely with other cars in real time;
the 2015 DARPA Robotics Challenge, created to develop semi-autonomous ground robots able to handle intricate tasks in hazardous, human-built environments; 
and, in 2025, the world's first humanoid-robot kickboxing tournament showcased the impressive hardware capabilities of modern humanoid robots.
Despite these, successful commercial applications of robotics are primarily found in car manufacturing, with little to no autonomous capabilities and relatively small domestic applications of autonomous robot vacuums and hotel delivery robots.

Although robot hardware has advanced in recent years, the robot's ability to reason effectively in real-world environments remains a significant challenge.
Machine learning solutions are generally considered by many to be the most promising avenue to resolve this issue. 
Therefore, many recent works focus on the integration of machine learning methods into robotic workflows.
A key contributing factor to the success of machine learning is the abundance of Python-based high-quality open-source software.
On the other hand, software for robotics is far more complex, requiring in-depth knowledge in several fields.

Robotics software has undergone significant evolution over the past several decades, yet it remains considerably more complex and fragmented than in fields such as machine learning.
To understand the current challenges in robotics software development, 
it is useful to consider how robotics software has evolved. 
Early industrial robots in the 1960s were programmed via record-and-playback mechanisms, with no real software abstraction~\cite{nilsson1984shakey, wave1972paul}. 
Around the same time, academic robotics began exploring software-controlled systems: Shakey~\cite{nilsson1984shakey}, for example, integrated symbolic AI planning, hierarchical control, and sensor feedback using LISP-based programs. 
The 1970s and 1980s saw the introduction of some of the first high-level robot programming languages (e.g., WAVE, AL, and RAPT)~\cite{wave1972paul, rapt1978ambler}, along with real-time embedded systems that enabled robots to respond to their environments. 
These developments improved modularity and sensing integration but were still limited by hardware specificity and lack of standardization.

In the 1990s, hybrid architectures combining deliberative planning and reactive control became more common, supported by libraries like the Robotics Toolbox~\cite{corke2002robotics}. This laid the groundwork for the 2000s emergence of middleware such as Player/Stage~\cite{gerkey2003player} and eventually ROS~\cite{quigley2009ros}, which standardized communication and hardware abstraction across research platforms. Despite these advances, robotics software today remains complex and fragmented—often requiring extensive C++ programming, specialized hardware drivers, and manual integration of learning components. 
Unlike machine learning frameworks, robotics lacks unified, Python-first tools that seamlessly support data collection, policy training, simulation, and deployment. 

Over the past few decades, 
artificial intelligence has seen several significant advances, such as 
the advent of deep learning~\cite{LeCun2015, Rumelhart1986}  
development of architectures such as convolutional and recurrent neural networks~\cite{NIPS2012_c399862d, lstm}, and 
more recently a class of models known as transformers have enabled many impressive results in a variety of fields such as image and natural language generation ~\cite{NIPS2017_3f5ee243, gpt4, deepseekr1}.
Due to the successes in other fields deep-learning has also shifted robotics research away from traditional model-based approaches~\cite{Khatib87, Posa14} and towards robot learning~\cite{Cheng24, pi0}; see reviews by 
Kroemer et al.~\cite{Kroemer21}, 
Xiao et al.~\cite{xiao2025robot},
Billard et al.~\cite{Billard2008}.\\

High-quality, well-documented software frameworks--PyTorch~\cite{NEURIPS2019_bdbca288}, Scikit-learn~\cite{scikit-learn}, OpenAI Gym~\cite{gym} and TensorFlow~\cite{tensorflow2015}--are one of the primary factors contributing to the success of machine learning: they are routinely used to teach university courses, drive the bulk of AI research output and power large-scale commercial products such as ChatGPT. 
With only a modest laptop, a newcomer can train their first neural network in under an hour by following the PyTorch quick-start tutorial, yet the same library also scales to the world's most advanced models~\cite{gpt4, deepseekr1}. 
Although all of these frameworks can run on a CPU, serious work usually demands specialized GPU hardware, and transferring data between devices is typically as simple as changing a single argument (e.g., the \texttt{device} flag in PyTorch).
The seeming ease of swapping hardware, however, hides a crucial caveat: in practice the target hardware device is almost always an NVIDIA GPU running CUDA and cuDNN. 
Alternative back-ends exist but remain less mature, and the ubiquity of the CUDA software–hardware stack has become so pronounced that many research projects and industrial pipelines now assume its presence by default~\cite{Dally21}. 
This de-facto standardization has accelerated progress by letting researchers write portable code and cloud providers offer uniform, highly optimized GPU instances at scale~\cite{chetlur2014, abadi2016}.

In contrast to the relative ease of developing and deploying machine learning applications, the development and deployment of robotics software is significantly more challenging for a variety of reasons--though this list is by no means exhaustive.
(C1)~Popular software frameworks for robotics (e.g., ROS and Orocos) require
steep learning curves for novice users, generally due to lack of documentation~\cite{Canelas23}, and are not readily integrated with tools for robot learning (e.g., data collection and processing, model training, and deployment).
(C2)~Unlike most machine learning packages--where you can build your application using entirely Python--robotics development still typically requires knowledge of C and C++. 
Most foundational robotics libraries exist only in those languages, exposing their functionality from Python requires skill with binding tools such as pybind11, Cython, or SWIG (among others).
(C3)~Robotics involves integrating a diverse range of hardware components--such as actuators, sensors, wheels, and onboard computers--each of which typically requires a custom driver interface or specific communication protocols, making it difficult to set up and switch between high-fidelity virtual simulations of custom robot setups.
(C4)~Developing robot systems require large teams with knowledge in many different fields such as 
control theory, 
kinematics and dynamics,
motion planning, 
computer vision and perception, 
signal processing, 
machine learning, and 
electrical and mechanical engineering~\cite{Kolak20}.

To address these challenges, we present \textsc{Ark}\footnote{Code available at \href{}{https://robotics-ark.github.io/ark\_robotics.github.io}.}---a Python-based robotics framework that is designed to accelerate development and prototyping, and enable users to deploy new methods on both simulated and real world robots. 
\textsc{Ark} is designed to integrate effortlessly with standard machine-learning workflows: it lets you gather data from simulators or real robots, preprocess it, and train policies with state-of-the-art imitation-learning methods such as ACT and Diffusion Policy. 
The main user-interface is designed based on the OpenAI Gym to make it familiar for machine learning researchers, and also so it integrates with common machine learning workflows (e.g., imitation learning). 
Additionally, we use a client-based system where various Python nodes can communicate over a network using a publisher-subscriber architecture.
Although \textsc{Ark} is designed as a Python-first framework, it also ships with utilities for exposing C/C++ functionality when performance matters.
Comprehensive documentation and several examples show how \textsc{Ark} can be installed, setup, and deployed in simulated and real robot setups. 
For advanced users, \textsc{Ark} exposes native ROS bindings, allowing seamless integration with existing ROS codebases.
Beyond the core environment interface, Ark provides reusable modules for low-level control, data collection, visualization, system identification, and mobile-base navigation. 
Several real-world and simulated case studies illustrate Ark's flexibility and ease of use.
In this paper, we provide an overview of our proposed framework called Ark, outlining the design principles that shaped its development and detailing its core capabilities. 
We then demonstrate Ark's versatility through extensive use-cases that give details on how to setup several examples and demonstrate the ease in which one can switch between simulation and the real robot system, 
various data-collection strategies for imitation learning, 
policy training with several established imitation-learning algorithms, 
map building via SLAM and motion planning for mobile robots, and 
finally we provide several embodied AI demonstrations.


\section*{Framework Overview}

\begin{figure}
    \centering
    \includegraphics[width=1\linewidth]{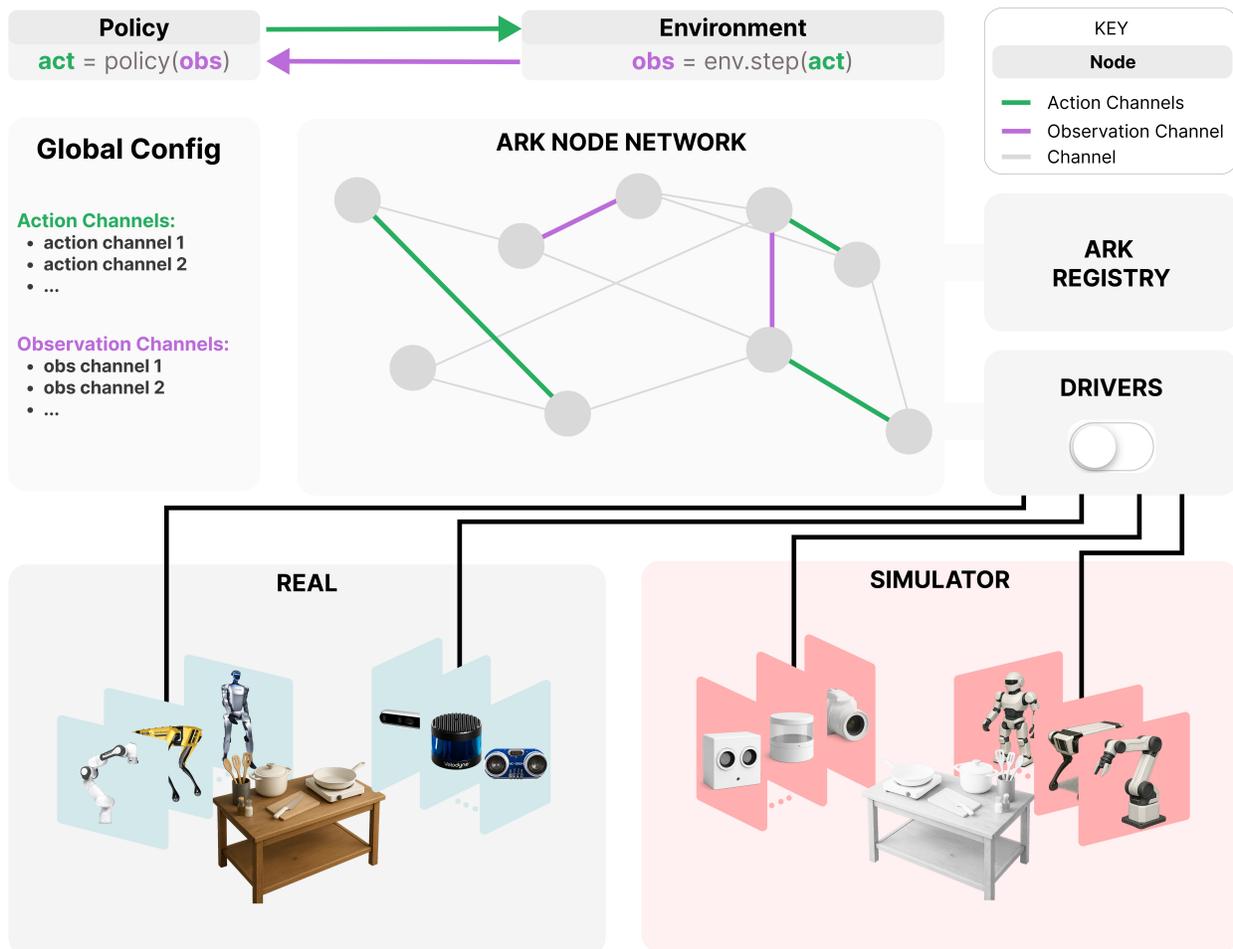}
    \caption{Ark uses a unified configuration file to define action and observation channels, which are then instantiated within a distributed node-based network. This architecture supports both real and simulated hardware through interchangeable drivers and identical communication interfaces. The Ark Registry manages active nodes, while each component (e.g., sensors, actuators, policies) operates as an independent process. As a result, pipelines developed in simulation can also be used on physical systems without code modification, ensuring a consistent sim-real interface.}
    \label{fig:ark-framework}
\end{figure}

Machine learning, based on deep learning, has emerged as one of the most promising and effective approaches to enable intelligent robots to complete complex reasoning tasks.
Although numerous studies report encouraging demonstrations, these systems have yet to progress beyond controlled laboratory settings into real-world deployments. 
A key limiting factor in this line of research is that there lacks a clear consensus on what constitutes an appropriate system architecture for embodied AI~\cite{roy2021}.
So far, recent works have largely converged on three alternative architectures.
The first approach assumes a library of parameterized skills executable by the robot, with an LLM or VLM selecting the appropriate skill at each environment step~\cite{mower2024ros, lan25}.
A second, trains a VLA model typically by fine-tuning a VLM to output directly actions that are executed on the robot~\cite{pi0}.
A third, uses a VLA or VLM to output action tokens in some latent space which are then mapped by another (often smaller) model to robot control actions~\cite{li2025hamster}.
Moreover, for each architecture setup, there also exist many problems around training these models (e.g, scaling data collection, sim-to-real, generalization) and deploying the results on real hardware.
In order to promote future research in embodied AI, 
Ark has been designed in such a way that is aligned with typical machine learning workflows and enables users to easily prototype novel architectures and deploy them on physical robots. 

We have implemented Ark with three core design philosophies in mind.
(D1) We design Ark's user interface to align with well-known machine learning libraries.
Robotics inherently demands expertise from a wide range of disciplines. With the growing influence of machine learning in the field, many researchers and engineers specializing in machine learning are now focusing on its deployment in robotic systems. 
However, robotics software still lacks the maturity and standardization seen in machine learning libraries, making development and adoption more difficult. 
Ark aims to bridge this gap by offering a more familiar and accessible interface for those coming from a machine learning background.
(D2) Developing and testing novel methods on real robots introduces a range of safety concerns; particularly when developers are in close proximity to the physical systems, as is commonly the case in robotics laboratories worldwide. 
While simulators provide a valuable environment for early-stage development, they often differ significantly in architecture from real-time, event-driven robotic systems. 
As a result, transitioning from simulation to physical deployment can be cumbersome and error-prone. 
Ark is designed to reduce this barrier by enabling seamless switching between simulated and real-world environments.
(D3) Python is arguably the most used programming language in machine learning, and it is widely regarded as significantly more user-friendly than languages like C, C++, or Java. 
In addition, Python boasts a vast ecosystem of well-maintained libraries and packages, making it an ideal choice for rapid development and experimentation.
We therefore have developed Ark with a Python-centric focus.
We acknowledge that in some cases, robot software is required to operate at high frequencies that is challenging for Python to handle (e.g., low-level controllers for dynamic tasks such as locomotion). 
In these cases, we offer a range of tools that help users to expose C/C++ code to Python.

The key aspects of the Ark framework are shown diagrammatically in Figure~\ref{fig:ark-framework}.
The following sub-sections provide an overview of the main features of Ark.

\subsection*{Ark Network}

A fundamental software design principle is modularity, which promotes maintainability, code re-use, and fault isolation~\cite{Parnas72}.
A robot system can be divided into specialized tasks such as data acquisition, state estimation, task planning, and control. 
The system's modules must communicate and exchange information to achieve these tasks. 
Modern operating systems facilitate this by enabling each module to run as a separate software process and communicate across a network, either on the same machine or distributed across different devices. 
Modularity and inter-process communication has been a long-standing practice in robotic software design~\cite{Montemerlo03, quigley2009ros, ros2}.

\paragraph{Message channels}

Ark uses a publisher-subscriber asynchronous message passing system for interprocess communication.
In the Ark Network (see the central part of Figure~\ref{fig:ark-framework}), each node represents a Python script running in its own process. 
These nodes are each given a unique name and communicate with one another through messaging channels.
Messages are transmitted across channels, allowing nodes to exchange information within the network.

The Ark Network is created programatically, each Python script implements a class that inherits from the \texttt{BaseNode} class.
Communication between nodes is established by calling the \texttt{create\_publisher} and \texttt{create\_subscriber} methods.
The user can choose an appropriate name for the messaging channel and define it with a string.
The full channel name is then given by \texttt{NODE\_NAME/CHANNEL\_NAME} where \texttt{NODE\_NAME} is the name of the node and \texttt{CHANNEL\_NAME} is given by the user.

The Ark Network low-level communication model is designed to be modular, allowing different networking libraries to be easily swapped out.
Currently, Ark uses the Lightweight Communications and Marshalling (LCM) library~\cite{Huang10} as its backend for network communication.
LCM is a middleware system with support for multiple programming languages.
We chose LCM for its lightweight design and its built-in tools for data collection, debugging, and introspection.

The modularity of our implementation of networking infrastructure is valuable, as it allows us to easily adapt the low-level networking layer to suit different design goals in the future, depending on research directions pursued by us or the broader community.
In future versions of Ark, we plan to support distributed training and potentially inference for backpropagation-based models, enabling more advanced machine learning workflows across networked nodes; 
such a goal can not be easily and efficiently handled by LCM in its current implementation. 

We follow the LCM type specification language to define message types; i.e., each message channel is defined by a name and a message type.
We provide a library called \texttt{ark\_types} containing many message types common to robotics (e.g., a \texttt{joint\_state\_t} type or a \texttt{transform\_t} type).

Another advantage of using LCM is its ability to facilitate integration of low-level languages such as C, C++, or Java into the Ark framework. 
Since nodes in Ark communicate via LCM messaging channels, scripts written in alternative languages can interact by utilizing the standard LCM publisher/subscriber interface to communicate over the network.
This approach can be useful in scenarios involving hardware devices--such as haptic interfaces--that are only accessible through C/C++/Java APIs provided by the manufacturer. 
In such cases, the user can expose the device to Ark by implementing the appropriate LCM publishers and subscribers.
However, due to Ark's architectural design for coordinating between simulation and real-world environments (discussed later), using LCM as a bridge between alternative languages and Python may not always be the best choice. 
To address this, and recognizing that C/C++ are the dominant languages in hardware development, Ark also provides a set of tools and helper functions/classes to assist users in directly exposing C/C++ functionality to Python.

\paragraph{Services}

Asynchronous communication is not always suitable for all tasks. 
To address this, Ark provides a request-response mechanism known as services. 
This pattern ensures a clear association between requests and responses, making it ideal for operations that require acknowledgment--for example, triggering a calibration routine on a robotic arm.
Ark services use the LCM message type specification for request and response types, allowing them to be selected dynamically.
Similar to message channels, Ark services can be identified by a name, chosen by the user to suit the desired task.

In order to connect services over the network, Ark includes a central registry, that acts as a lightweight coordination and discovery hub.
Information about the network can be retrieved (e.g., active services from other nodes) and enables various features such as runtime visualization and fault isolation.
The registry has default service names, these are identified in list of services by the \texttt{\_\_DEFAULT\_SERVICE}.
The Ark registry must be run by the user before the Ark Network can be established by other nodes.

\paragraph{Launcher}

As illustrated in Figure~\ref{fig:ark-framework}, multiple nodes can be executed and connected to form the overall Ark Network. 
Each node can be launched individually by starting its corresponding Python script from the terminal. 
However, manually launching a large number of processes can become cumbersome and error-prone.
To address this, Ark provides a launcher utility that allows users to define the entire Ark Network within a single configuration file using the YAML format. 
This launcher script can then be executed once from the terminal to automatically start all specified subprocesses, simplifying the initialization of complex network setups.

\subsection*{Observation and Action Channels}

We adopt the standard terminology from the reinforcement learning literature, referring to \textit{observations} and \textit{actions} as follows. 
An \textit{observation} is information received from the environment, typically taken from sensors such as cameras or joint encoders. 
An \textit{action} is a control signal or decision variable that influences the environment, for example by specifying joint velocities sent to a robot's actuators.

To promote ease of use and reduce the learning curve for users familiar with machine learning, Ark provides an interface inspired by the well-known OpenAI Gym (now Gymnasium) library. 
An environment class implements a \texttt{reset} function, which returns an observation and an information dictionary. 
Also a \texttt{step} function is given that accepts an action as input and returns the next observation, a reward, termination and truncation flags, and an information dictionary.

We define observation and action spaces each by a collection of message channels running on the Ark Network.
Each space is defined in the constructor of the environment class using a dictionary which maps the message channel name to it's type.
Observation and action space classes are initialized as class attributes for the environment class that both automatically handle communication with the Ark network:
the observation space subscribes to each channel, whereas the action space publishes to each given channel.
By allowing the user to choose the observation/action spaces enables them to easily prototype different input/outputs for their policy model architecture. 
Also, each channel in the observation space, may be published at different sampling frequencies.
Each observation returned thus contains the most recent message recieved on each channel.

\subsection*{Real World and Physics Simulation}
\begin{figure}
    \centering
    \includegraphics[trim={2em 3em 2em 10em}, clip, width=1\linewidth]{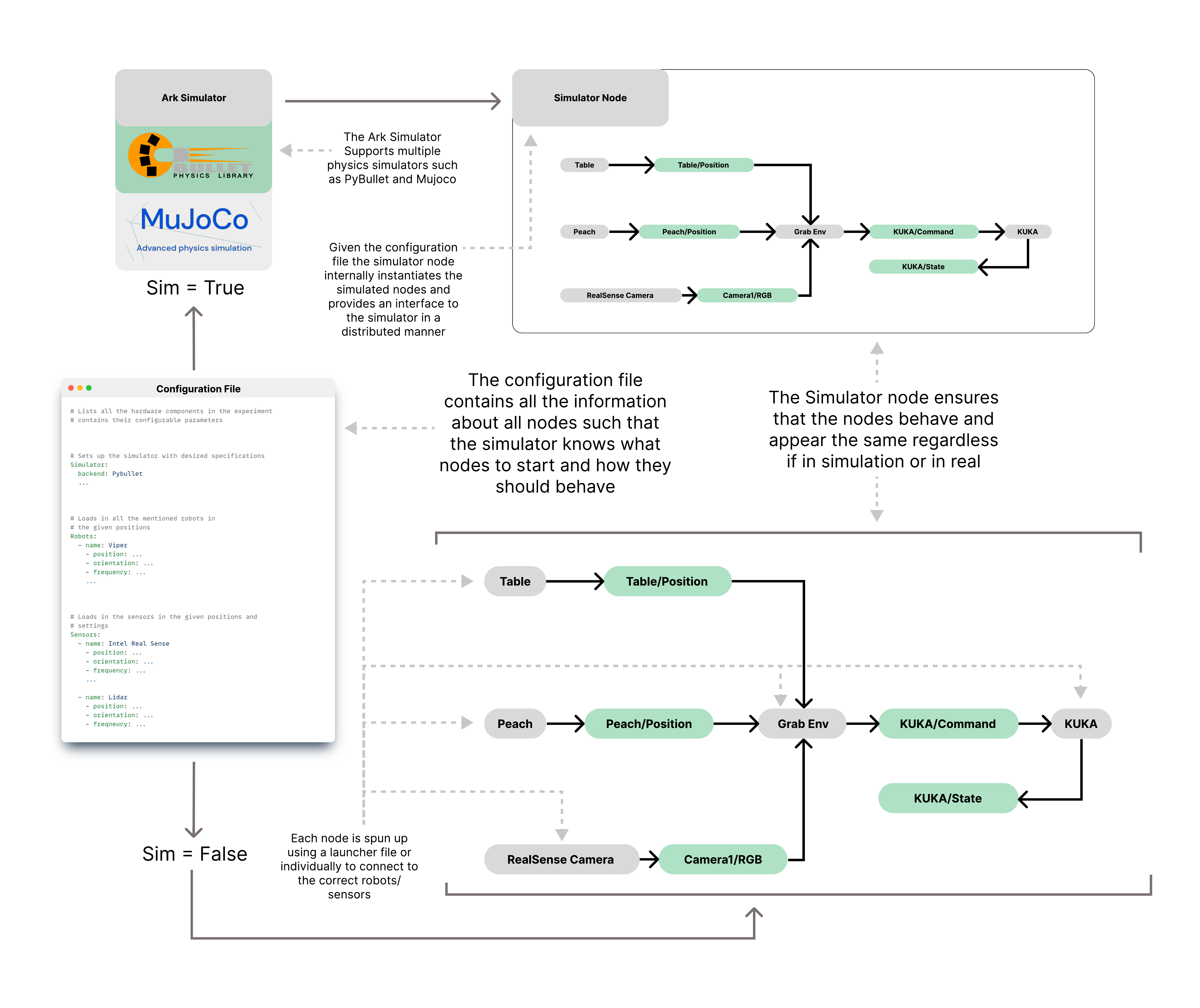}
    \caption{Technical diagram illustrating how Ark uses a unified configuration file to instantiate a distributed simulated system that mirrors real-world deployments. The YAML-based configuration specifies robots, sensors, environments, and networking parameters, which the Ark Simulator parses to launch corresponding simulated nodes. Each component, such as robot controllers, cameras, and sensor emulators, runs as an independent process, communicating through the same message passing protocol used in real deployments. This ensures that policies and pipelines developed in simulation operate identically when transferred to physical hardware, facilitating seamless sim-to-real transitions and reproducible experimentation}
    \label{fig:sim_real_tehcnical}
\end{figure}

As outlined in our three core design philosophies, the second principle (D2) focuses on ensuring that Ark can seamlessly operate across both real and simulated environments. 
This section provides further details on how this is achieved, including an overview of how Ark interfaces with various simulators.

\paragraph{Sim-Real switch}

A key capability of \textit{Ark} is its ability to easily switch between simulated environments and real-world robotic systems using a single configuration flag, i.e., \texttt{sim = True, False}. 
This is made possible by Ark's distributed, node-based architecture, where each robot and sensor--whether physical or simulated--is implemented as an independent computational node. 
Ark distributes the physics simulators, using a configuration file as seen in Figure ~\ref{fig:sim_real_tehcnical}, which takes the chosen simulator and internally spins up nodes to mimic the interface from a real system. 
This ensures consistency across deployments and enabling transparent switching between them. 
Further details on hardware drivers are provided in the following section.

\paragraph{Simulator backend}

There are many physics simulators available to the robotics community, however there does not exist a single simulator that performs best in all desired features for all domains~\cite{Collins21}.
Each simulator is typically developed
for a specific purpose or with a certain application in mind, e.g. manipulation, medical, marine, soft robotics, locomotion, etc.
Therefore, instead of directly interfacing with a single simulator, we instead provide a simulator backend which enables users to integrate, in theory, any simulator they wish that fits their need. 
Currently, Ark supports  PyBullet and MuJoCo due to their popularity in machine learning.
In future work, we plan to integrate IssacSim, and will consider suggestions from the community.

The choice of backend used, as well as the switch between simulated and physical systems, is entirely managed through Ark's configuration infrastructure. 
By modifying a single YAML file, users can define the desired setup (e.g., real/sim and if sim then which simulator), and Ark will automatically initialize the appropriate drivers, ensuring consistent message schemas, channel names, and execution flow.

\subsection*{Robot and Sensor Drivers}
\label{Robot and Sensor Drivers}

While Ark provides some structure in its user interface (i.e., Gym-like interface as discussed above), it is very extensible and we have designed Ark with broad comparability across robots and sensors in mind.
Recently developed frameworks such as LeRobot~\cite{cadene2024lerobot} and PyRobot~\cite{pyrobot2019}, target only specific robot embodiments, Ark on the other hand is design to ensure that a wide range of hardware can be easily integrated. 
We enable generalizability across hardware by providing several interfacing methods. 

\paragraph{Python drivers}

We provide an abstract Python base class, \texttt{ComponentDriver}, designed to standardize the integration of hardware components with Ark. 
To implement a driver, users create a subclass and override standard abstract methods such as \texttt{get\_data} (for sensors) and \texttt{send\_command} (for robots). 
Each driver integrates with Ark's global sim-real switch, automatically routing messages to either real or simulated hardware based on the global configuration settings.

\paragraph{C++ drivers}

As mentioned above, programming in C/C++ is sometimes necessary in robotic systems. 
For instance, certain hardware components provide only C/C++ interfaces, and in some cases, high sampling frequencies are required for real-time performance, such as in locomotion control. 
To support these scenarios, we provide a set of tools written in C++ using the \texttt{pybind11} library, enabling users to expose their hardware components to Ark. 
These tools ensure that hardware components with only C++ interfaces can be integrated with Ark in a consistent manner, following the same conventions as Python-based drivers.

\paragraph{ROS-Ark driver}

Arguably, ROS is the most widely used robotics platform today, with many research groups and developers around the world relying on it to build their robotic systems. 
In fact, some robots (e.g., ViperX arms) are only supported via manufacturer-provided ROS interfaces. 
To enable integration with the ROS ecosystem, Ark includes a ROS-Ark driver that facilitates bidirectional communication between ROS topics and Ark message channels. 
This allows users to run existing ROS setups while simultaneously leveraging Ark's interface, without requiring any modifications to the original ROS codebase. 
Moreover, the driver serves as a migration tool, helping users transition their existing ROS-based systems to Ark.
Currently, based on our experience that most research labs (including our own) continue to use ROS 1, the ROS-Ark bridge supports only ROS 1. 
Support for ROS 2 may be considered in the future, depending on user demand.

\subsection*{Tools for Introspection and Debugging}

\begin{figure}
    \centering
    \includegraphics[width=1\linewidth]{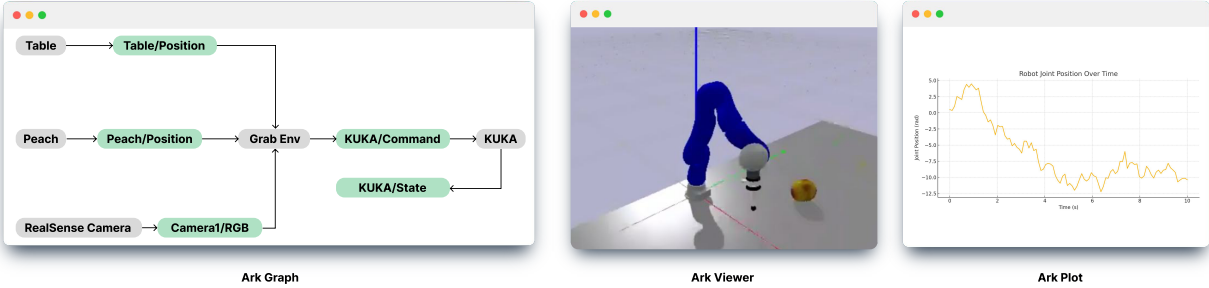}
    \caption{ Graphical debugging tools provided by Ark. Ark Graph displays active nodes and their communication channels for network analysis. Ark Viewer renders live image streams to support camera calibration and inspection. Ark Plot visualizes real-time numerical data on any channel, aiding in system monitoring and debugging.}
    \label{fig:ark-tools}
\end{figure}

Robot systems are often complex and rely on many intercommunicating processes.  
As a result, having a comprehensive suite of debugging tools, visualized in Figure ~\ref{fig:ark-tools}, is essential, enabling users to efficiently investigate and resolve issues as they arise.

\paragraph{Ark Graph}

The Ark Graph tool offers a real-time visual representation of all active nodes, their published and subscribed channels, and available services.

\paragraph{Ark Plot}

Ark Plot is a real-time plotting tool for visualizing numeric data on Ark message channels. 
It allows users to monitor the evolution of variables over time, making it useful for tasks such as tuning control parameters, diagnosing sensor behavior, and debugging system performance.

\paragraph{Ark Viewer}

Ark Viewer enables real-time visualization of image data transmitted over any LCM channel, making it especially valuable for configuring, monitoring, and debugging camera systems.

\paragraph{LCM Tools}

Another reason LCM was chosen as the communication library is the suite of built-in tools it offers for debugging and introspection. 
For example, \texttt{lcm-spy} is a graphical tool for viewing messages on an LCM network. 
Similar to network analysis tools like Ethereal/Wireshark or \texttt{tcpdump}, it allows users to inspect all received LCM messages and provides detailed information and statistics on the channels in use (e.g., number of messages received, message Rate in Hz, and jitter in ms).

\section*{Use Cases}

In this section, we provide an overview of several use-cases for our proposed framework.
These use-cases illustrate the ease of using and setting up Ark for multiple situations common to robot learning.
All code to reproduce these use-cases will be made available.

\subsection*{Switching Between Simulation and Reality}


\begin{figure}
    \centering
    \includegraphics[width=1\linewidth]{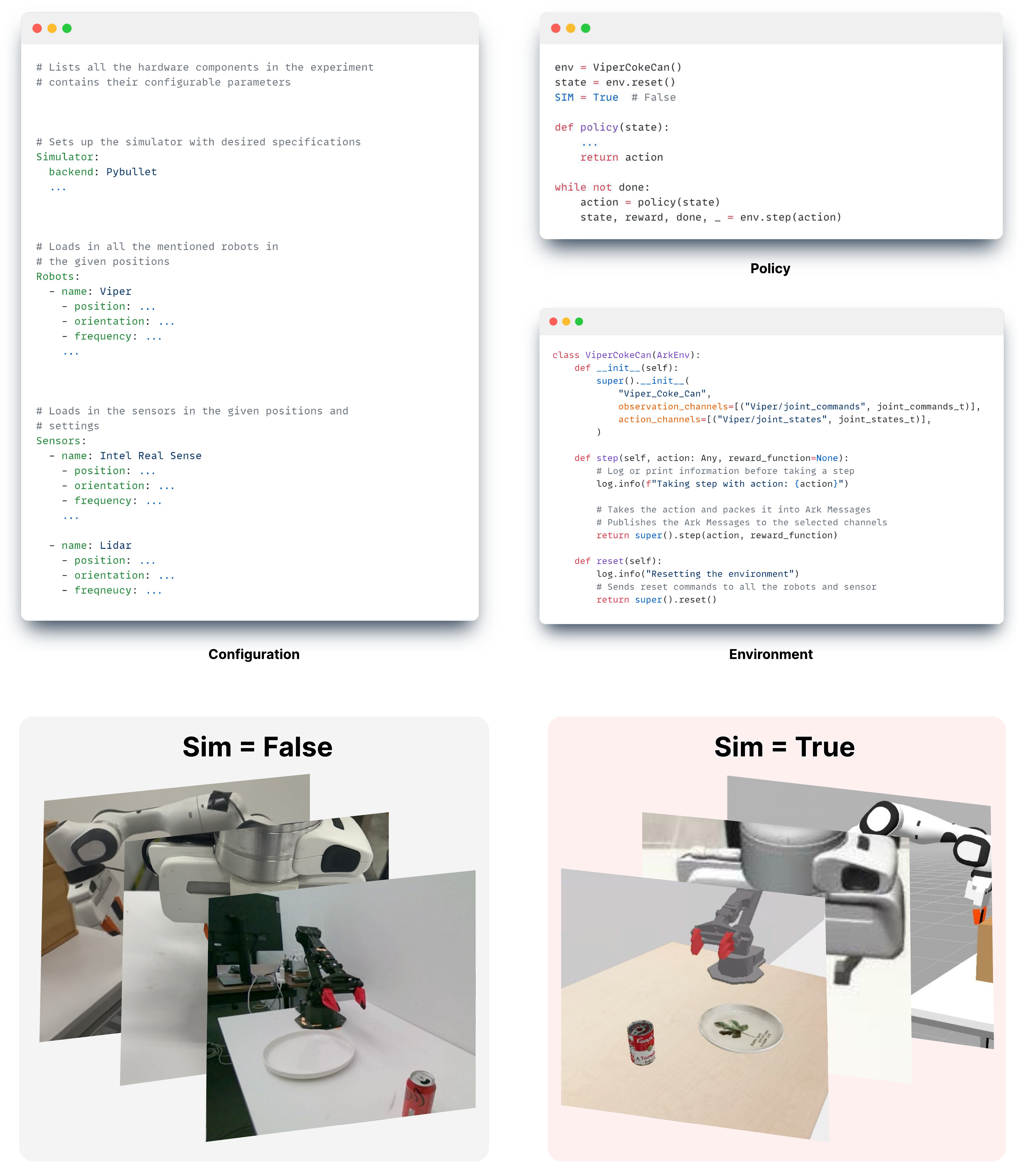}
    \caption{Seamless sim-real transition in Ark enabled by consistent observation and action space definitions. The environment configuration specifies sensor inputs (e.g., joint states, images) and actuator outputs (e.g., joint commands), which remain identical across both simulated and real systems. By toggling a single flag (sim = True/False), Ark automatically routes data through the different observation and action channels, allowing a single policy pipeline implementation to operate in both domains without modification. This unified interface simplifies development, debugging, and deployment across the sim-real boundary.}
    \label{fig:sim-real}
\end{figure}

Deploying learned robot policies in real-world scenarios can raise safety concerns.
Moreover, many existing frameworks lack straightforward implementation pathways, resulting in adhoc and inconsistent solutions that do not generalize well across different systems or embodiments.

Ark addresses these challenges by providing a unified control interface built on Python/C++ drivers. 
It offers a configurable abstraction layer that enables seamless deployment of robotic policies across both simulated and physical environments.
After defining the environment (i.e, robot, sensors, objects), the user need only specify a single flag in the configuration to switch between real and simulation, i.e., \texttt{sim = True, False}.

To demonstrate this facility, we implemented a  pick-and-place task using a ViperX 300s fixed-base robotic arm equipped with a parallel gripper at the end-effector. 
The robot was tasked with picking an object and placing it onto a plate. 
The entire environment is specified by a single YAML file that defines object initial placements, camera and robot poses, as well as physics parameters such as gravity.
This configuration is used by Ark to setup the physical environment (left part of Figure~\ref{fig:sim-real}), and instantiate a simulation environment where the scene geometry  mirrors the real-world setup (right part of Figure~\ref{fig:sim-real}).

In this example, the observation space is the current joint position command, and the action space is the goal robot joint velocity command.
A hand-crafted expert policy is used in both the real world and simulation. 

Crucially, transitioning the same policy to the real robot required no changes to code or data structures--only the single configuration variable \texttt{sim} was toggled. 
Ark internally reroutes communication from the simulated drivers to the physical hardware while preserving the same observation and action channels. 
This abstraction ensures that all downstream code (e.g., the policy logic, environment wrappers, and logging infrastructure) remains unchanged, facilitating direct deployment and reproducibility.

\subsection*{Data Collection for Imitation Learning}

\begin{figure}
    \centering
    \includegraphics[width=1\linewidth]{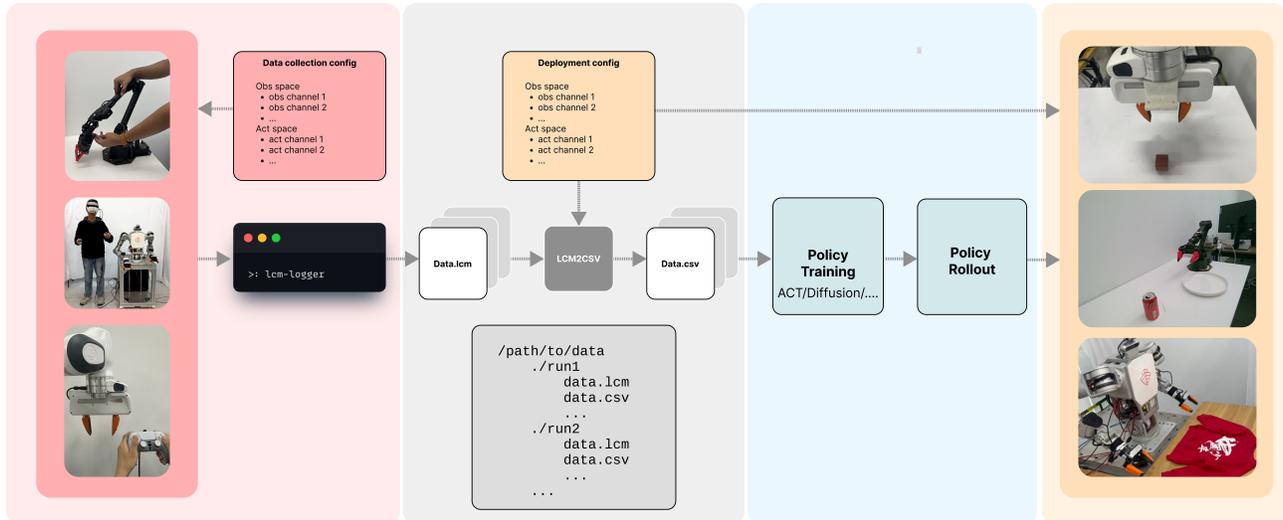}
    \caption{lcm-logger enables efficient data collection for imitation learning by recording demonstrations from a variety of control interfaces, including kinesthetic teaching, VR teleoperation, and gamepad input (left). Each demonstrations is saved as a separate CSV file (right), allowing users to accumulate diverse datasets across different input modalities rapidly.}
    \label{fig:data_collection}
\end{figure}

Learning policies using imitation learning requires large demonstration datasets.
Data collection typically requires humans to interact with the robot system to collect several demonstrations of a collection of tasks~\cite{billard2008survey}.

Two primary methods can be employed to collect data from the system: \emph{kinesthetic teaching} and \emph{teleoperation}. 
In kinesthetic teaching, the human physically interacts with the robot to provide demonstrations (see the left part of Figure~\ref{fig:data_collection}). 
This approach is generally considered intuitive for human operators~\cite{muxfeldt2014kinesthetic} and is often preferred~\cite{li2025}; however, it may raise safety concerns due to the need for direct physical contact with the system.
The second method, teleoperation, involves the human controlling the robot remotely via an interface such as a virtual reality headset and controllers or a gamepad (central and right parts in Figure~\ref{fig:data_collection}). 
Teleoperation allows the operator to interact with the robot from a safe distance, but it introduces challenges such as limited visibility~\cite{Gleicher19} and difficulty in mapping controller inputs to the robot's control dimensions~\cite{Mower19}. 
As a result, effective teleoperation often requires a skilled operator~\cite{mower2022optimization}.

Due to Ark's modular architecture and the use of distinct message types for each communication channel, setting up data collection pipelines is straightforward. 
This design enables users to easily swap out different interfaces as needed.
LCM provides a utility called \texttt{lcm-logger}, which can be executed at any time to record data. It captures and writes all messages published on LCM channels to a log file.
Ark includes built-in functionality to extract data from these log files and convert it into CSV format. 
Moreover, by utilizing the same observation and action space configuration defined in the environment class, the data can be extracted in a format consistent with that used during deployment.
Since messages on different channels may be published at varying frequencies, Ark also provides tools to handle this asynchrony. 
These tools allow users to either interpolate the channel data or extract the most recent message at each time step.

\subsubsection*{Kinesthetic Teaching}

Ark supports learning from human-guided demonstrations through kinesthetic teaching and also replay.
Instead of deploying a pre-programmed expert policy, the ViperX 300s robotic arm is manually guided through the task by the human physically interacting with the system.
During this process, the LCM logger records the channel data, capturing the full trajectory of the demonstration as it is executed.
We also include a camera in the setup, whilst this is not used in the policy for this demonstration, it can be used to collect videos of the robot setup.
Note, the observation and action spaces remain the same as in the previous section.

Once a demonstration is complete, the data is saved to a log file.
This data can either be processed for use in training an imitation learning policy, or replayed on the system using the \texttt{lcm-logplayer}. This replay functionality allows the demonstration to be reproduced exactly as it was performed by the human demonstrator.
Such playback is particularly useful in scenarios where the human appears in the camera frame--potentially introducing bias or noise into vision-based policies--or when force-based interactions are involved. 
In the latter case, it can be difficult to disentangle the forces that the robot should actively track from those resulting from physical contact during the demonstration.
Additionally, we provide services to streamline environment resetting. 
For instance, users can link a reset service to a keyboard button press, making repeated demonstrations or evaluations more efficient and user-friendly.

\subsubsection*{Teleoperation}

Ark also supports teleoperation using input devices such as VR and gamepad controllers.
In one setup (central part of Figure~\ref{fig:data_collection}), a user controls the OpenPyRo-A1 humanoid robot by streaming, in real-time, 6-DoF poses from a VR controller over a local network.
Using an inverse kinematic controller, these are converted to joint velocity goals that are sent to the robot actuators.
A second setup (right part of Figure~\ref{fig:data_collection}) has a similar architecture for the Ark Network, however in this case a PlayStation4 controller is used as the interface that controls the robot gripper pose.

\subsection*{Imitation Learning}

In this section, we demonstrate several use-cases for implementing imitation learning.
These use-cases focus on how to use Ark for data collection, training, and deployment.
Also, we highlight the effectiveness of using the OpenAI Gym (Gymnasium) interface with Ark to allow easy adaptations to the policy.
An overview of Ark's data collection nodes are shown in Figure~\ref{fig:data_collection}.

We showcase two methods for imitation learning:
(i) diffusion policy,
(ii) action chunking with transformers, and
Ark serves as the central infrastructure, providing modular components, standardized interfaces, and real-time communication, all of which streamline data collection and deployment of trained policies.


\subsubsection*{Diffusion Policy}

\begin{figure}
    \centering
    \includegraphics[width=1\linewidth]{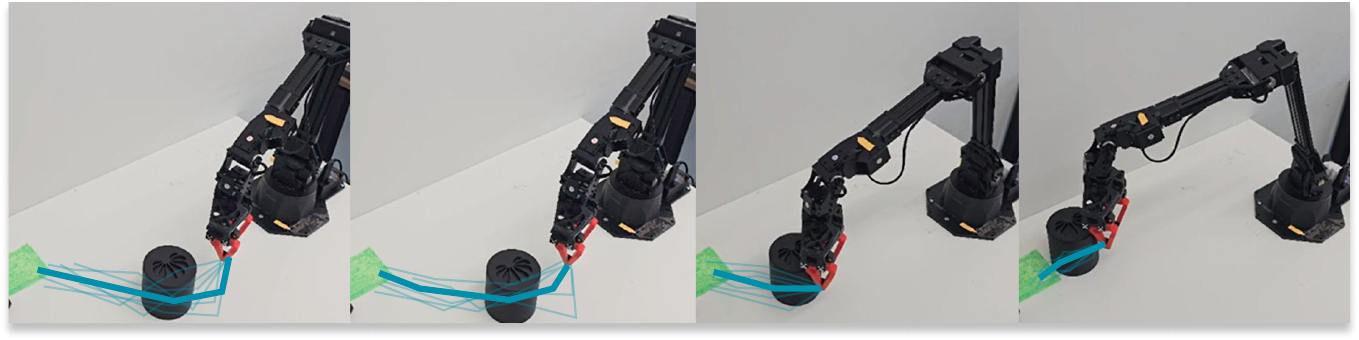}
    \caption{Sequential snapshots of a Viper X 300 s arm executing a learned diffusion policy. The line overlay shows the sampled action trajectories converging toward the target object, while the arm autonomously refines its actions at each timestep.}
    \label{fig:diff-policy}
\end{figure}

First, we demonstrate deployment of a learned diffusion policy~\cite{Cheng24}.
For this use case, a ViperX300s robotic arm and an Intel RealSense RGB camera are configured to perform a pushing task, shown in Figure~\ref{fig:diff-policy}.
The observation space includes the robot's joint positions and a continuous RGB image stream from the camera.
The action space consists of joint position commands sent to the ViperX 300s.

Data collection is facilitated entirely through Ark nodes as demonstrated in the previous section, which operate as independent, reusable processes.
In this setup, four key nodes are deployed: a Controller Node for PS4 joystick input, an Environment Node that translates joystick commands into target end-effector poses, an Inverse Kinematics Node that converts these poses into joint commands, and Sensor Nodes that publish RGB images and joint states.
Because Ark enforces strict message typing and channel separation via LCM, users can modify or replace nodes (e.g., switch cameras or use a scripted controller) without rewriting the rest of the system.
This modularity enables reuse across tasks, which is particularly beneficial for research workflows where hardware configurations can change or various interfaces can be swapped out for different contexts.

After each demonstration, the environment is required to be reset.
When the user presses the `X' button on the controller, the trajectory is saved automatically, and the robot resets to a neutral state.
This mechanism is built directly into Ark's service framework, enabling resets without requiring custom scripting.
As a result, users can collect high-quality demonstration datasets with minimal setup, and without needing to manage state transitions or converting data formats manually.

Once the data is collected, Ark already specifies the observation and action space each as a collection of channels. 
This same configuration is used to extract data form the log files making it easy to implement data loaders in imitation learning training scripts.

Deployment within Ark replicates the data collection setup, easing engineering complexity.
The trained diffusion policy is loaded into a policy node within the same environment used for data collection, replacing the joystick controller node.
The policy receives the current RGB image and robot joint state from the observation channels and outputs the target end-effector position. 
This target position is published as a command through the same action channels used during demonstrations.
Since the execution pathway remains unchanged, users do not need to modify the underlying infrastructure to test learned policies, Ark ensures a consistent system setup across training and deployment.

\subsubsection*{Action Chunking with Transformers}

\begin{figure}
    \centering
    \includegraphics[width=1\linewidth]{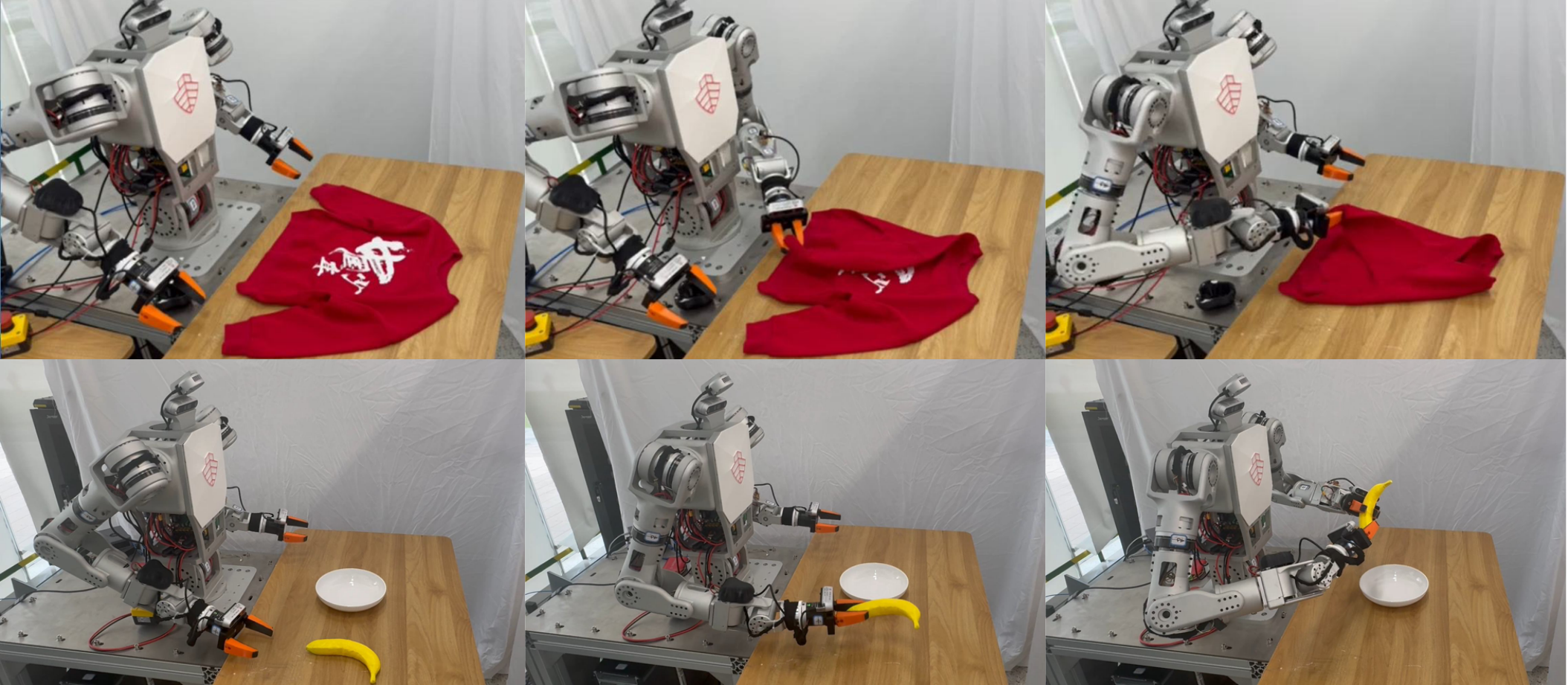}
    \caption{Execution of an ACT-based policy on the OpenPyro humanoid robot for two distinct tasks: cloth manipulation (top row) and object handover (bottom row). The policy produces precise, contact-rich behaviors that enable the robot to flatten a shirt and place a banana into a bowl.}
    \label{fig:Pyro-ACT}
\end{figure}

In addition to diffusion policy, we have also implemented a demonstration based on action chunking with transformers~\cite{zhao2023}.
For this demonstration, we use the OpenPyro-A1 humanoid platform~\cite{huang2025openpyro} and teleoperate it with a virtual reality headset.
Data was collected for two tasks, the first is cloth manipulation task (top row of Figure~\ref{fig:Pyro-ACT}) and the second is object handover task (bottom row of Figure~\ref{fig:Pyro-ACT}); snapshots in Figure~\ref{fig:Pyro-ACT} are from the videos of the robot performing these tasks using the learned policies.





\subsection*{Mobile-base Robots}

Many real-world tasks, such as inspection, require a robot to navigate to various locations within an environment. 
To do this effectively, the robot must be able to both access a map of the environment and determine its position within that map.
This process is commonly known as Simultaneous Localization and Mapping (SLAM), 
which enables the robot to construct a map of an unknown environment while simultaneously estimating its own location.
Once a map has been created and the robot is accurately localized, planning algorithms can be employed to navigate the environment.

We have developed a mobile robotics pipeline in Ark that implements a variant of FastSLAM~\cite{thrun04}. 
Using teleoperation to control the robot and data from its onboard LIDAR sensor,
we employ FastSLAM to construct a map of the robot's environment. 
Once the map is built, we apply the A* search algorithm for path planning, incorporating a distance transform to maintain safe navigation margins between the robot and nearby obstacles. 
A proportional-derivative (PD) controller then guides the robot through each waypoint, translating the planned trajectory into low-level wheel velocity commands.

\subsubsection*{Map Building}
\begin{figure}
    \centering
    \includegraphics[width=1\linewidth]{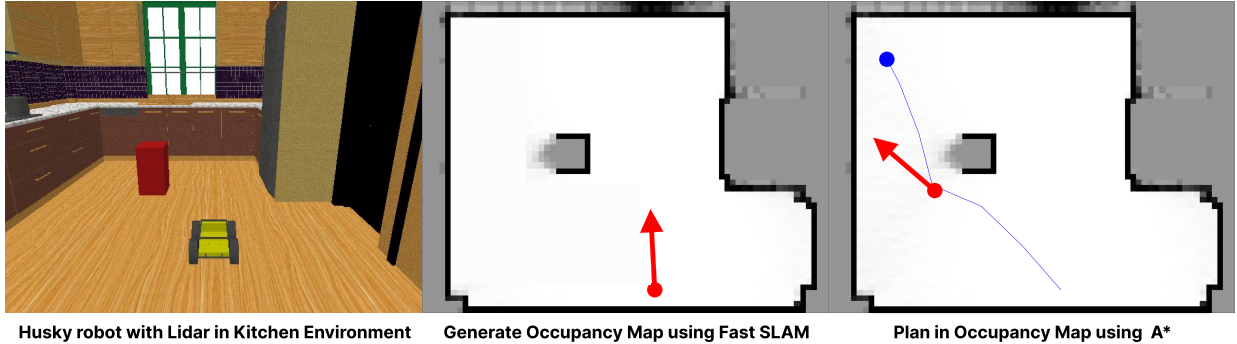}
    \caption{Using Ark's integrated SLAM and visualization tools, a Husky robot equipped with a LiDAR sensor navigates a kitchen environment. The robot first constructs an occupancy map using FastSLAM (center), facilitated by Ark's modular data streaming and map-building nodes. The final panel (right) shows an A* path generated within the occupancy map using Ark's planning and rendering utilities. This setup demonstrates how Ark enables end-to-end navigation workflows—from sensor integration and SLAM to path planning and visualization—within a unified framework.}
    \label{fig:map-building}
\end{figure}

To construct a 2D map of the environment while estimating the robot's pose, a teleoperation assisted SLAM pipeline was implemented using the Ark framework. 
The pipeline consists of two primary subsystems: 
a teleoperation controller and a probabilistic SLAM module, both implemented as Ark nodes that communicate over message channels.

The teleoperation node allows a user to control the robot by specifying desired linear and angular velocities. 
These high level commands are published over an Ark action channel and received by the low level controller node, which computes the corresponding left and right wheel velocities using differential drive kinematics. 

Simultaneously, the LIDAR data and control signals are streamed to the SLAM node, which fuses these inputs to estimate the robot's pose and construct a map of the environment. 
The LIDAR data are represented as $N\times 2$ arrays of polar coordinates (distance, angle) centered at the LiDAR frame, while the control signals reflect the robot's commanded motion. These data are consumed by a node running FastSLAM, implemented using a Rao Blackwellized Particle Filter. 
Each particle maintains both a pose estimate and a local occupancy grid, with cells assigned probabilities between 0 (free) and 1 (occupied).
Figure~\ref{fig:map-building} illustrates Ark's integration with SLAM for map building and navigation.

\subsubsection*{Navigation}


Given a map of the environment and the robot's pose within that map, the system can perform motion planning while avoiding obstacles. 
The A* search algorithm, a widely used grid-based planning method, has been adapted for robotic motion planning and is implemented in Ark, integrated with the map generated via SLAM. 
A* is particularly appealing because it is a complete algorithm--meaning that if a valid path from the start to the goal exists, A* is guaranteed to find it, provided the environment is discretized finely enough.

Upon receiving a goal position in world coordinates, the planning subsystem is triggered. This node receives as input both the target location and the probabilistic occupancy map produced during SLAM. 
To plan a safe and efficient trajectory, the planner first discretizes the occupancy map using a predefined threshold (typically 0.5) to distinguish between free and occupied cells.
It then computes the distance transform of the grid to determine, for each cell, the shortest distance to the nearest obstacle. 
This metric is used to ensure that planned paths maintain a minimum clearance of at least half the robot's width plus a user-specified margin, reducing the risk of collision in narrow passages.
The shortest feasible path from the robot's current location to the goal is computed using the A* algorithm. 
The resulting trajectory is a sequence of Cartesian waypoints $(x,y)$ that avoid obstacles while maintaining smooth curvature and safety margins. 
To improve execution efficiency, the raw path is then downsampled based on a user-defined spatial resolution, reducing unnecessary intermediate waypoints and promoting smoother motion.

The control subsystem consumes both the planned path and the real-time pose estimates from the SLAM node. 
Implemented as a PD controller, the system directs the robot through each waypoint in order. For each waypoint, the controller first rotates the robot to face the target using angular control, followed by linear control to advance toward it. 
Once the current waypoint is reached within a defined tolerance, the next target is selected. 
The controller outputs both linear and angular velocities, which are then converted into left and right wheel commands using differential drive kinematics.

\subsection*{Embodied AI}


\begin{figure}
    \centering
    \includegraphics[width=1\linewidth]{images/figures/deepseek.pdf}
    \caption{Deep Seek integration with Ark Framework to allow the Viper to play board games}
    \label{fig:DeepSeek-Ark}
\end{figure}

Large language models (LLMs) and vision-language models (VLMs) have been shown to endow robots with impressive reasoning capabilities~\cite{mower2024ros, lan25}.
Ark's modular design and Python-first architecture make it well suited for integrating large language models (LLMs) and vision-language models (VLMs) as high-level policy selectors within robot control loops; Python is the most common programming language used in machine learning.
An agentic system was implemented using the Viper robotic arm, in which high-level semantic reasoning is performed by an LLM to perform reasoning tasks, shown in Figure~\ref{fig:DeepSeek-Ark}.
As our base LLM we use Deepseek-R1~\cite{deepseekr1}.
The system implemented follows a code-as-policy paradigm~\cite{liang2023code}: each robotic manipulation skill--such as ``pick piece'', ``place at location'', and ``remove object''--are implemented as a parameterized policy that is callable as a Python function. 
These functions are exposed through a policy library, which the LLM selects from based on task context and scene understanding in the prompt.

Each node in the Ark Network--scene perception, language-based reasoning, and motion execution--is encapsulated as a standalone node. 
DeepSeek is integrated as an Ark node that exposes a service interface; the prompt is the request input and the output of the LLM is returned as the service response. 
At each step of the policy is a decision event, the perception node publishes scene observations (e.g., board state, object poses, RGB images) to a shared channel. 
These inputs are then passed as a structured query to the DeepSeek node via an Ark service call.
The query includes the current state of the environment, a list of available policy functions, and optionally, a task prompt expressed in natural language.
We tested Deepseek-R1 versus other LLMs (Qwen 2.5 and Llama 3) in a round-robin tournament and found DeepSeek-R1 to have the highest win-rate: Qwen 2.5 (26.6\%), Llama 3 (30.0\%), and
Deepseek-R1 (43.3\%).
Unfortunately, each LLM was unable to beat humans.




\section*{Discussion}

Ark has been purpose-built from the ground up with a design philosophy that will feel familiar to those experienced with machine learning software.
It is guided by a number of core principles: simplicity and modularity, Python-first accessibility, and seamless integration with both real-world and simulated robotic systems. 
With its flexible, distributed architecture and lightweight communication layer, Ark enables reliable coordination of sensors, actuators, and AI models across a wide range of robotic embodiments. 
It offers researchers and developers a robust and customizable framework for rapidly prototyping, deploying, and iterating on real-world robotic systems.

Through a series of use-cases, Ark has been validated on several tasks important to robot learning research; from dexterous manipulation with expert-coded policies to language-conditioned visuomotor control powered by state-of-the-art foundation models. 
These implementations demonstrate Ark's role not only as an interface, but as a catalyst for advancing research in embodied AI. 
Its unified environment abstraction, standardized data pipelines, and native support for machine learning workflows enable teams to transition smoothly from simulation to real-world hardware, and from prototyping to full deployment.

\subsection*{Related Work}

Over the years, several robotics frameworks have been developed to address different aspects of robot control, each with varying levels of modularity, language support, and real-time capabilities. 
We provide a comparison between these different robotic frameworks can be seen in Table~\ref{tab:compare}. 
The comparison points in the table cover several points in each column. 
\textit{Python}: indicates if Python is a supported language.
\textit{C/C++}: indicates if C/C++ are supported languages.
\textit{Gym}: indicates if the main user interface is designed to align with OpenAI Gym (Gymnasium).
\textit{IL}: means if imitation learning algorithms are directly integrated into the framework.
\textit{RL}: means if reinforcement learning algorithms are directly integrated into the framework (whilst Ark currently does not support RL, in the future we plan to implement this functionality).
\textit{Sim}: indicates whether simulators are integrated in the framework.
\textit{ROS Dep}: indicates if ROS is a dependency of the framework, we see this as a negative attribute.
\textit{ROS Con}: indicates whether an optional ROS connection is provided (e.g., in our case, we provide the ROS-Ark Bridge).
\textit{PubSub}: indicates if the framework implements a modular publisher-subscriber networking approach.
\textit{Sim-Real}: indicates if specific switching mechanisms exist to switch between simulation and reality.
\textit{Pip}: indicates whether the framework is installable using only \texttt{pip}.
\textit{Data tools}: indicates whether tools are provided off-the-shelf for data collection/processing.
\textit{Inspection tools}: indicates whether tools are provided off-the-shelf for inspection and debugging.
\textit{Limited Embod}: indicates if the framework limits the user to a number of specific embodiments, this is a negative point.

YARP is a peer-to-peer communication framework focused on modularity and performance, primarily used in humanoid and legged robotics like iCub and MIT Cheetah. 
However, its ecosystem is limited by exclusive support for C++. 

LCM, on its own, offers a lightweight publish-subscribe model optimized for low-latency, high-bandwidth messaging and has broad language support. 
While highly effective for communication, it provides only the messaging layer, requiring additional infrastructure for tasks such as system coordination, simulation integration, and machine learning workflows--gaps that Ark addresses.

OROCOS provides real-time libraries for control, kinematics, and filtering, with strong deterministic guarantees via CORBA integration--but it leaves much of the broader system design to the user.

ROS 1, once the de facto standard for robotics development and now officially end-of-life, offered a rich ecosystem of reusable tools, libraries, and drivers that greatly accelerated prototyping and system integration. 
However, it also suffered from several architectural limitations, including a lack of built-in security and poor reliability over lossy networks. 
Since Ark is currently focused on research rather than commercial deployment, security is not a primary concern at this stage.
Additionally, switching between real and simulated environments in ROS 1 was not seamless and often required users to implement custom, ad hoc solutions.

In contrast to the frameworks/libraries mentioned above, Ark is designed to support modern machine learning-driven robotics. 
It promotes modularity and reusability but removes unnecessary complexity by providing a Python-first, lightweight, and distributed architecture. 
Unlike frameworks like YARP or OROCOS that focus on specific layers of the stack, Ark offers end-to-end integration--from low-level hardware communication to high-level policy control--using standardized channels and services. 
Its minimal setup and direct compatibility with machine learning tools make it particularly well-suited for rapid iteration and embodied AI research.
Moreover, installing Ark is simple, it can be setup using pip and is compatible with conda environments. 

\subsection*{Future Work}

To better support the needs of embodied AI research, future development of Ark will focus on two key areas: reinforcement learning (RL) infrastructure and high-fidelity simulation capabilities.

While RL is a popular approach in modern robotics, Ark's current infrastructure offers only limited support for RL workflows. 
Upcoming enhancements will include native integration with popular RL libraries such as StableBaselines3 and RLlib, as well as support for parallelized environment execution. 
These improvements will enable researchers to train, evaluate, and deploy RL policies efficiently across both simulated and physical robotic platforms using a unified environment abstraction.

In parallel, we aim to significantly advance Ark's simulation stack. 
Although Ark currently supports integration with simulators like PyBullet and MuJoCo, it lacks advanced features such as domain randomization and differentiable physics--both crucial for robust policy learning and sim-to-real transfer. 
Future releases will focus on tighter integration with high-performance simulation backends, enabling more accurate, scalable, and versatile simulations.

\subsection*{Epilogue}

In summary, Ark represents a significant step forward in bridging robotics and machine learning through a modern, modular, and accessible software architecture. 
By lowering the technical barriers to real-world robot deployment, while maintaining flexibility and extensibility for advanced research, Ark empowers a new generation of researchers to develop, test, and deploy intelligent robotic systems more efficiently. 
As it continues to mature--through enhanced simulation support, deeper RL integration, and expanded tooling--Ark is well-positioned to serve as a framework for embodied AI, catalyzing progress across the robot learning community.

\begin{table}[t]
\centering
\caption{Comparison of Ark versus alternatives. Note, red mark indicates a negative feature.}
\label{tab:compare}
\begin{tabular}{l|llllllllllllll}
           & \rotatebox{90}{Python} 
           & \rotatebox{90}{C/C++} 
           & \rotatebox{90}{Gym} 
           & \rotatebox{90}{IL} 
           & \rotatebox{90}{RL} 
           & \rotatebox{90}{Sim} 
           & \rotatebox{90}{ROS Dep} 
           & \rotatebox{90}{ROS Con} 
           & \rotatebox{90}{PubSub} 
           & \rotatebox{90}{Sim-Real} 
           & \rotatebox{90}{Pip} 
           & \rotatebox{90}{Data tools}
           & \rotatebox{90}{Inspect. tools} 
           & \rotatebox{90}{Limited Embod.} \\ \hline
\textbf{Ark} & \mark & \mark & \mark & \mark &  &\mark &       & \mark & \mark & \mark & \mark & \mark & \mark &       \\
LeRobot    & \mark &       & \mark & \mark & \mark &\mark &       &       &       & \mark & \mark & \mark &       & \textcolor{red}{\mark} \\
PyRobot    & \mark &       & \mark &       &       &\mark & \textcolor{red}{\mark} & \mark &  & \mark &       &       &       & \textcolor{red}{\mark} \\
ROS 2      & \mark & \mark &       &       &       &\mark &   NA  &  NA   & \mark &       &       &       & \mark &       \\
Orocos     &       & \mark &       &       &       &      &       &       & \mark &       &       &       &       &       \\
YARP       &       & \mark &       &       &       &      &       &       & \mark &       &       &       &       &       
\end{tabular}
\end{table}

\clearpage
\section*{References}

\makeatletter
\renewcommand{\section}[2]{}  
\makeatother

\bibliography{bib}




\newpage

\appendix

\end{document}